%% file: main.tex
\documentclass[conference]{IEEEtran}
\IEEEoverridecommandlockouts
\usepackage{cite}
\usepackage{amsmath,amssymb,amsfonts}
\usepackage{algorithmic}
\usepackage{graphicx}
\usepackage{caption}
\usepackage{subfig}
\usepackage{enumitem}
\usepackage{amsmath}
\usepackage[margin=1in]{geometry} % Ajusta as margens
\usepackage{textcomp}
\usepackage{xcolor}
\usepackage{hyperref}

\usepackage{tikz}
\definecolor{lime}{HTML}{A6CE39}
\DeclareRobustCommand{\orcidicon}{%
    \begin{tikzpicture}
    \draw[lime, fill=lime] (0,0) circle [radius=0.16] 
    node[white] {{\fontfamily{qag}\selectfont \tiny ID}};
    \draw[white, fill=white] (-0.0625,0.095) circle [radius=0.007];
    \end{tikzpicture}
    \hspace{-2mm}
}
\newcommand{\orcidID}[1]{\href{https://orcid.org/#1}{\orcidicon}}
\def\BibTeX{{\rm B\kern-.05em{\sc i\kern-.025em b}\kern-.08em
    T\kern-.1667em\lower.7ex\hbox{E}\kern-.125emX}}
\begin{document}

\title{Instance space analysis of the capacitated vehicle routing problem\\
%\thanks{ São Paulo Research Foundation (FAPESP), Brazil, Proc. 2022/16276-4.}
}

% --- 3. SUBSTITUA TODO O SEU BLOCO \author{...} POR ESTE ---
% --- SUBSTITUA SEU BLOCO \author POR ESTE PARA CENTRALIZAR ---
\author{
    \IEEEauthorblockN{
        Alessandra M. M. M. Gouvêa\textsuperscript{1}\orcidID{0000-0002-0512-0162},
        Nuno Paulos\textsuperscript{1}\orcidID{0009-0008-0524-3913},
        Eduardo Uchoa\textsuperscript{2}\orcidID{0000-0002-8687-2613}, Mariá C. V. Nascimento\textsuperscript{1}\orcidID{0000-0002-3094-6847}
    }
    \IEEEauthorblockA{
        \textit{\textsuperscript{1}Divisão de Ciências da Computação, Instituto Tecnológico de Aeronáutica (ITA), São José dos Campos, SP, Brazil}\\
        \textit{\textsuperscript{2}Departamento de Engenharia de Produção, Universidade Federal Fluminense (UFF), Niterói, RJ, Brazil}\\
        % Opcional: E-mails podem ser adicionados aqui se necessário, mas mantenha a clareza.
    }
}

\maketitle
\begin{abstract}
This paper seeks to advance CVRP research by addressing the challenge of understanding the nuanced relationships between instance characteristics and metaheuristic (MH) performance. We present Instance Space Analysis (ISA) as a valuable tool that allows for a new perspective on the field. By combining the ISA methodology with a dataset from the DIMACS 12th Implementation Challenge on Vehicle Routing, our research enabled the identification of 23 relevant instance characteristics. Our use of the PRELIM, SIFTED, and PILOT stages, which employ dimensionality reduction and machine learning methods, allowed us to create a two-dimensional projection of the instance space to understand how the structure of instances affect the behavior of MHs. A key contribution of our work is that we provide a projection matrix, which makes it straightforward to incorporate new instances into this analysis and allows for a new method for instance analysis in the CVRP field.
\end{abstract}

\begin{IEEEkeywords}
capacitated vehicle routing problem, instance space analysis, metaheuristic
\end{IEEEkeywords}

%AQUI
\input{tex01_Indroduction}
\input{tex02_RelatedWork}

\input{tex04_Methodology}
\input{tex05_Results}

\input{tex06_conclusion}

\section*{Acknowledgment}
The authors are also grateful for the financial support provided by FAPESP (2022/16276-4, 2022/05803-3, 2013/07375-0) and CNPq (403735/2021-1; 309385/2021-0).

\bibliographystyle{IEEEtran}
\bibliography{mybib}

\vspace{12pt}

\end{document}

%% file: tex01_Indroduction.tex
\section{Introduction}
The Vehicle Routing Problem (VRP) is a combinatorial optimization problem widely studied in the literature \cite{laporte1992}. The goal is to find a minimum-cost set of routes for a fleet of vehicles while meeting the customers' demands and operational constraints. The most classic and highly investigated variant of the VRP is the Capacitated VRP (CVRP), where the fleet is homogenous and the only restriction on the routes is that the sum of the served demands does not exceed the capacity of a vehicle.
The literature on metaheuristics (MHs) for the CVRP is vast, since tailored MHs offer good solutions with reduced computational cost, especially for large-scale instances \cite{maximo2024ails,vidal2020concise}. The literature recognizes, with strong consensus, that no MH can offer ideal performance across all instances of a particular optimization problem. Consequently, it is natural that researchers seek to determine what makes an MH perform well for a particular problem instance, which is directly related to the characteristics of instances that may impact MH performance \cite{smith2009cross}.

Understanding what drives MH behavior enables a more reliable comparison of algorithms and facilitates the identification of regions of the instance space in need of new benchmarks \cite{smith2023instance}. The comparative analysis of metaheuristics commonly relies on previously established benchmark instances. However, failing to prioritize instance selection can produce a biased set of instances, favoring certain MHs. For this reason, a proper consideration of the heterogeneity of instances is essential to guarantee a more robust comparative analysis. Additionally, the need for an optimal set of instances can reveal the limitations of the current benchmarks and guide the creation of new instances. Despite these arguments, few studies have focused on these issues in the VRP literature \cite{rasku2016feature,rasku2019feature,karkkainen2020application}.

Issues found in traditional comparative studies --— in particular, the use of too homogeneous benchmarks and analyses based on median performance --— are not unique to the field of VRP. To address the shortcomings, Kate Smith and co-workers developed a series of published works that led to a methodology named Instance Space Analysis (ISA) \cite{smith2023instance}. ISA emerges in the literature as a promising alternative approach that focuses on a different way to evaluate algorithms. The ISA methodology seeks to build a comprehensive view of the set of all possible instances of a problem. By describing instances through their features and applying dimensionality reduction and machine learning techniques to map each instance into a 2D space (named instance space), ISA shifts the focus of algorithm evaluation to the visual exploration of the relationship between instances, algorithms, and their characteristics. By providing tools for visualization, analysis, and generation of test instances, ISA opens the path for a more reliable evaluation of algorithms and a more insightful understanding of their performance based on the instances characteristics.

This paper presents an investigation of the CVRP instance space by evaluating the relationship instance-algorithm performance on a set of algorithms designed for the CVRP through ISA methodology. The used data was collected from the DIMACS 12th Implementation Challenge on Vehicle Routing. The provided data regards instances and results from the algorithms that participated in the competition. To perform the evaluation, we employed the ISA framework \cite{smith2023instance} which is based on statistical analysis and machine learning techniques.

The main contributions of this paper are:

\begin{itemize}
    \item  We unveil relationships between instance characteristics and the algorithm performance data extracted from the literature.
    \item We investigate how node distribution/clustering, route topology (distances, levels of connection), demand, and vehicle capacity influence CVRP algorithm performance, measured by the Primal Integral \cite{berthold2013measuring}.
    \item We transform raw data (instances and algorithm performance) into knowledge using the PRELIM, SIFTED, and PILOT stages.
    \item Our findings include the identification of relevant features and 2D-instance space visualization, thus facilitating the understanding of different CVRP instances.
\end{itemize}

The remainder of this paper reviews the literature on instance characterization and algorithm selection for VRP/TSP in Section \ref{sec:related_work}. Section \ref{sec:methods} details the ISA methodology used. Section \ref{sec_results} presents the VRP instance space analysis results. Section \ref{sec:conclusion} concludes and outlines avenues for future work.

%% file: tex02_RelatedWork.tex
\section{Related Work}
\label{sec:related_work}

The inherent complexity of combinatorial optimization problems has driven research beyond the mere search for optimal solutions. Studies have aimed to investigate the subtle nuances that distinguish more complex instances from less complex ones \cite{smith2012measuring}. Such works usually represent instances through feature vectors to map the characteristics that influence optimization difficulty \cite{smith2011discovering,mersmann2013novel}. Knowing the relationship between instance characteristics and the performance of solution methods enables the selection and configuration of algorithms with less reliance on human and computational resources \cite{rasku2019feature}, facilitates the generation of more robust benchmarks, such as more representative and challenging test instances \cite{uchoa2017new}, and contributes to a broader understanding of optimization challenges \cite{smith2012measuring}. Therefore, the generated knowledge is expected to facilitate the development of more effective, resilient, and better-suited solution techniques that address the diversity of real-world challenges \cite{karkkainen2020application}. 

In the literature on the VRP, Rasku and Musliu \cite{rasku2016feature} point out the scarcity of studies on VRP instance features and the widespread use of simplistic adaptations of TSP features, thereby neglecting the specific traits of the VRP. Building upon previous work, Rasku et. al. \cite{rasku2019feature} proposed and evaluated an extensive set of features for VRP instances, aiming to identify the most relevant ones, and identified ten, among more than four hundred, as promising. However, their analysis was restricted to heuristics, without exploring the potential of state-of-the-art MHs. 

The next section shows the contribution of this study to overcome the limitations mentioned in this section. For this, the research presented in this paper utilizes a subset of the features proposed by Rasku et. al. \cite{rasku2019feature}, but differently from these authors, we consider state-of-the-art MH, a larger set of instances and the application of ISA. The ISA methodology employed in our study provides a way to incorporate future instances into this analysis in a straightforward way, given its capacity to generate a projection matrix and to define the relevant instance characteristics.

%% file: tex04_Methodology.tex
\section{Methods}
\label{sec:methods}
This section presents the methodology employed for the ISA of the CVRP. All data employed relies on  the algorithms' performance on the instances tested in the first phase of DIMACS 12th Implementation Challenge on Vehicle Routing. The data is publicly accessible, whereas instances from the second phase remain undisclosed. 

\subsection{Collecting meta-data about instances}
\label{subsec:collecting}

The first step for ISA is to collect instance features -- mathematical and statistical measures describing instance characteristics -- and algorithm performance metrics on these instances. 

%ThAs mentioned b\textcolor{red}{Se eu fosse você, eu explicaria direto explicar a respeitoIn 2022, DIMACS hosted a vehicle routing implementation challenge. This competition assessed the practical performance of algorithms, departing from theoretical worst-case analysis. The DIMACS event brought together active research groups and produced a dataset of state-of-the-art algorithm performance across CVRP instances. Our work leverages this competitively generated DIMACS dataset of algorithm performance on standard CVRP instances. 

%The event had two phases; our analysis focuses on the first phase because only its CVRP instances are publicly accessible, whereas instances from the second phase remain unpublished.
The instances collected are diverse, including classic CVRPLib instances, such as E-n101-k8 and CMT4; a set of 100 systematically generated ``X'' instances with varied depot, customer, demand, and route size attributes; and 12  real-world instances from the companies Loggi and ORTEC, with distances based on road networks and driving times, which are now also available on CVRPLib.

A critical meta-data for ISA is the set of instance features. Previous studies, as discussed in Section \ref{sec:related_work}, have explored various features for characterizing TSP and VRP instances. We kept the organization of features into the six categories proposed in Rasku and Musliu \cite{rasku2016feature}. Indeed, the chosen features, shown next, are a subset of the features already considered in the literature. The reader may consult the indicated references for more details on each feature.

%Following Rasku and Musliu \cite{rasku2016feature}, we organized the features into distinct categories, each containing statistical data and measures of different natures. As the chosen features are a subset of those considered by them, we kept the categories However, the labels

%The labels of the instance features remain unchanged from the original work and we refer to their reference for a more detailed description of them.  The selected instance features are enumerated next, organized into distinct categories.}

\subsubsection{Node distribution (ND)}
The features in this category quantify and describe the spatial distribution of customers and the depot. They aim to address the question: ``How does the spatial arrangement of customers and the depot influence the difficulty of finding a high-quality solution?''. In our study, we consider the following ND features: 

%Distance matrix \cite{mersmann2013novel,kanda2011selection} (ND1)

\begin{enumerate}[start=1,label={\bfseries ND\arabic*:},leftmargin=*,labelindent=10pt]
\item Basic distance matrix statistics \cite{mersmann2013novel,kanda2011selection}: average, standard deviation, median, kurtosis, etc
\item Total of Edge lower cost \cite{kanda2011selection}
\item Fraction of distinct distances \cite{mersmann2013novel}
\item Position (x,y) of centroid \cite{smith2011discovering}
\item Distance of customers to the centroid \cite{smith2011discovering}
\item Number of clusters \cite{smith2011discovering,mersmann2013novel}
\item Size of clusters \cite{mersmann2013novel}
\item Distance of cluster centroids \cite{mersmann2013novel}
\item Ratio of the number of clusters to the number of cities \cite{smith2011discovering}
\end{enumerate}

\subsubsection{Minimum spanning tree (MST)}
The Minimum Spanning Tree (MST) of a graph connects all vertices with the lowest total edge cost. The following features capture aspects of the MST, aiming to answer the question: ``How does the structure of the minimum connectivity between nodes influence the difficulty of optimization?''. The following features are computed based on the MST:

\begin{enumerate}[start=1,label={\bfseries MST\arabic*:},leftmargin=*,labelindent=10pt]
\item Edge cost \cite{mersmann2013novel,hutter2014algorithm}
\item Node degree \cite{mersmann2013novel,hutter2014algorithm}
\item MST depth from the depot \cite{mersmann2013novel,hutter2014algorithm}
\end{enumerate}

\subsubsection{Probing features}
This category contains features derived from running an algorithm (heuristic or exact) on an instance within a limited time or number of steps. These features aim to answer the question: ``How does a specific algorithm perform when attempting to solve a problem under constrained time or effort, and what does this reveal about the instance's difficulty?''. In the literature, probing features are derived from local search heuristics \cite{pihera2014application} and branch-and-cut algorithms \cite{hutter2014algorithm}. Here, we analyze features based on the Lin-Kernighan heuristic, leaving branch-and-cut algorithms for future research. The following features are computed based on the Lin-Kernighan heuristic:

\begin{enumerate}[start=1,label={\bfseries P\arabic*:},leftmargin=*,labelindent=10pt]
\item Number of best improving steps \cite{pihera2014application,hutter2014algorithm}
\item Edge lengths in quartiles \cite{pihera2014application}
\item Tour segment length\cite{pihera2014application}
\item Edge count in tour segment\cite{pihera2014application}
\item Edge length in segment\cite{pihera2014application}
\item Tour cost from construction heuristic \cite{mersmann2013novel,hutter2014algorithm,pihera2014application}
\item Local minimum tour length \cite{pihera2014application,mersmann2013novel}
\item Tour intersections in plane \cite{pihera2014application}
\item Improvement per step \cite{mersmann2013novel,pihera2014application}
\item Steps to local minimum \cite{pihera2014application,hutter2014algorithm,mersmann2013novel}
\item Probability of edges in local minima \cite{mersmann2013novel,hutter2014algorithm}
\end{enumerate}

\subsubsection{Geometric features}
The features in this category analyze the spatial configuration of nodes, capturing shape information while abstracting from non-geometric characteristics like demands or capacities. These features aim to answer the question: ``What is the general shape of the problem and how does the arrangement of the nodes influence its optimization difficulty?''. The following features are used to analyze the geometry of the problem:

\begin{enumerate}[start=1,label={\bfseries G\arabic*:},leftmargin=*,labelindent=10pt]
\item Area of the enclosing rectangle \cite{smith2011discovering,mersmann2013novel}
\item Convex hull area \cite{mersmann2013novel,hutter2014algorithm}
\item Ratio of points on the hull \cite{mersmann2013novel,hutter2014algorithm}
\item Distance of enclosed points
to the convex hull contour \cite{mersmann2013novel}
\item Edge lengths of the convex hull
\end{enumerate}

\subsubsection{Nearest neighborhood (NN) features}
This features' category regards relationships between each node and its nearest neighbors, capturing the local search space structure and node connectivity. Unlike MST or geometric features, which focus on overall structure, NN features focus on relationships between a node and its closest nodes. These features aim to answer the question: ``How are nodes connected to each other in their immediate neighborhoods, and how do these connections influence the optimization difficulty?''. The following features are used to describe the neighborhood of each node:

\begin{enumerate}[start=1,label={\bfseries NN\arabic*:},leftmargin=*,labelindent=10pt]
\item Distance to 1st NN \cite{smith2011discovering,hutter2014algorithm}
\item Number of strongly connected components \cite{pihera2014application}
\item Number of weakly connected components \cite{pihera2014application}
\item Size of strongly connected components \cite{pihera2014application}
\item Size of weakly connected components \cite{pihera2014application}
\item Node input degree in directed
kNN graph \cite{pihera2014application}
\item Ratio of number of strongly and weakly connected
components \cite{pihera2014application}
\item Angles
between a node and its two nearest neighbor nodes \cite{mersmann2013novel}
\end{enumerate}
\subsubsection{VRP Specific Features}
This category of features is composed by values obtained directly from the parameter values of the VRP instances, such as vehicle capacity, customer demands, etc. Unlike the more generic features belonging to the previously described categories, these features regard  details which are specific to the VRP instance. These features aim to answer the question: ``How do customer demands, vehicle capacity, and other VRP constraints influence the difficulty of the problem and the quality of the solutions obtained?''. The following features are considered in this category:

\begin{enumerate}[start=1,label={\bfseries VRP\arabic*:},leftmargin=*,labelindent=10pt]
\item Distance from
centroid to the depot \cite{rasku2016feature}
\item Distance from customer to the depot \cite{rasku2016feature}
\item Client demands \cite{rasku2016feature}
\item Ratio of total demand to
total capacity \cite{rasku2016feature}
\item Number of customers
per vehicle \cite{rasku2016feature}
\end{enumerate}

\subsection{Performance of Algorithms}\label{subsec:algorithms}
\begin{figure*}[h!]
\centerline{\includegraphics[width=0.9\textwidth]{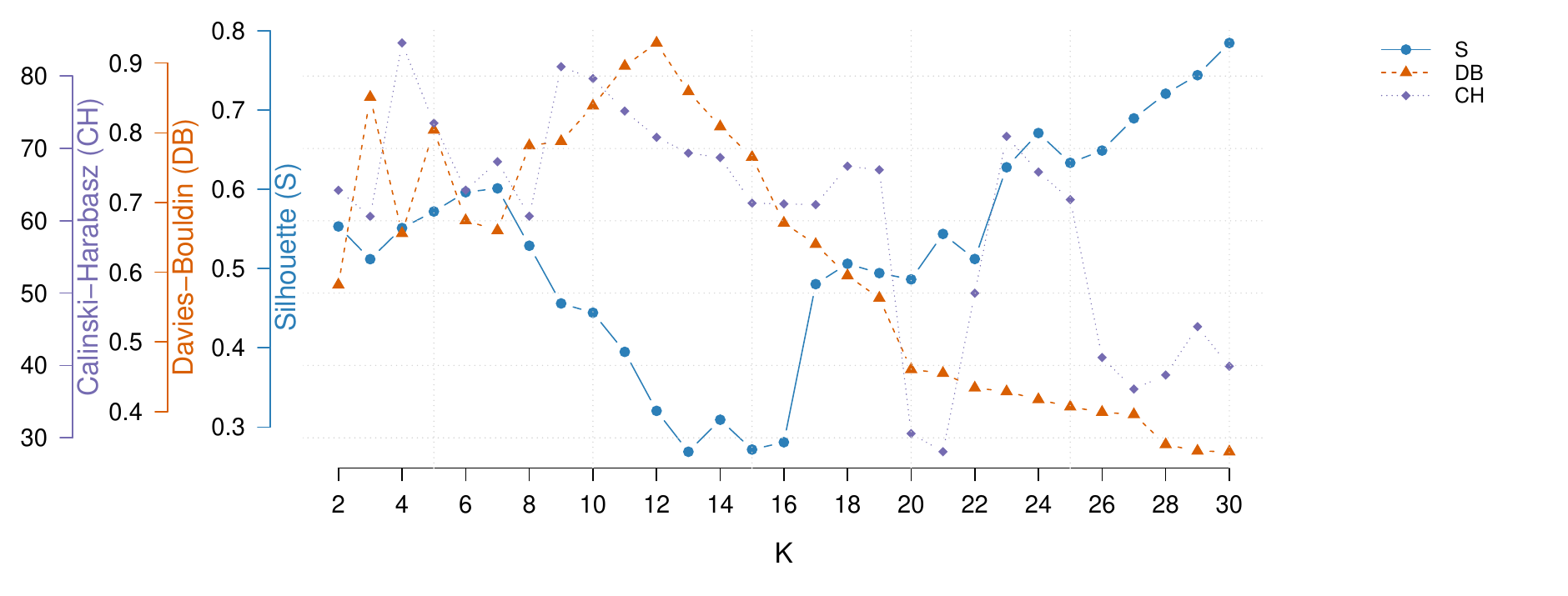}}
\caption{Analysis of the clustering metrics used to select a suitable K value in the SIFTED stage. The curves show the variation of Silhouette, Davies-Bouldin (DB), and Calinski-Harabasz (CH) metrics with different K values.}
\label{fig:varyingK}
\end{figure*}

The Primal Integral (PI) \cite{berthold2013measuring} was used as a performance metric, evaluating both solution quality and execution time; lower PI values indicate better performance. 
This information is important for projecting the instance data, since a step of the framework correlates each feature to the algorithm performance to keep features with high correlation with the performance of the algorithm. The set of algorithms is composed by the top eight finalists of the first phase. This investigation will not discuss the algorithm performance but only evaluate the instance space using such information from the algorithms.

\subsection{Constructing the Instance Space}
The construction of the instance space involves feature selection and dimensionality reduction. This process enables the identification of salient features and maps instances into a two-dimensional space while preserving their relationships, facilitating effective visualization. The ISA methodology achieves this through three interdependent methods applied sequentially: PRELIM, SIFTED, and PILOT. Next, we provide a brief explanation of these methods and their parameters.

\subsubsection{PRELIM}
PRELIM, short for Preparation for Learning of Instance Meta-data, standardizes and transforms algorithm characteristics and performance data, ensuring its suitability for subsequent steps. The method is configured by the following parameters:
\begin{itemize}
    \item $\mathbf{F}$: The matrix of instance characteristics (i.e., features), where each row represents a specific instance and each column represents a numerical feature describing that instance. $\mathbf{F}$ was computed as described in Section \ref{subsec:collecting};
    \item $\mathbf{Y}$: The matrix representing the measured performance of algorithms, where each row is associated with an instance, and each column stores the performance metric of a particular algorithm on that specific instance. $\mathbf{Y}$  contains the performance data produced by the DIMACS event as described in Section \ref{subsec:algorithms};
    \item $\epsilon$: The performance threshold, $\epsilon=0.15$, defines the lower bound for ``good'' performance, based on the high median of the Performance Indicator (PI) distribution for  the selected algorithms from the first phase of the DIMACS Challenge;
    \item $\phi_{\text{max}}$:  boolean flag that indicates whether the objective is to maximize or minimize the performance metric. We set $\phi_{\text{max}} = \text{false}$  because our objective is to minimize the performance metric;
    \item $\phi_{\text{bnd}}$ A boolean flag that indicates whether no limitation is applied to feature values to reduce the impact of outliers, in this investigation, $\phi_{\text{bnd}} = \text{false}$;
    \item $\phi_{\text{nrm}}$: A boolean flag that indicates whether feature and algorithm performances data were already normalized using MinMax normalization, therefore, we set $\phi_{\text{nrm}} = \text{false}$.
\end{itemize}

\subsubsection{SIFTED}
SIFTED, short for Selection of Instance Features to Explain Difficulty, identifies features that significantly influence algorithm performance. This is achieved in two steps. First, the absolute Pearson correlation between each feature and algorithm performance is computed. Features are selected if their correlation is greater than $0.5$ or less than $-0.5$. Subsequently, similar features are grouped using K-means clustering, where K represents the number of clusters. The dissimilarity measure is defined as $1-|p_{i,j}|$ represents the absolute Pearson correlation between features $i$ and $j$. To determine a suitable value for $K$, we analyzed the Silhouette, Davies-Bouldin (DB), and Calinski-Harabasz (CH) metrics, which evaluate cluster cohesion, separation, and variance. By varying K, we aimed to minimize Silhouette and DB while maximizing CH. Figure \ref{fig:varyingK} presents the metric curves; based on this analysis, we set $K=23$.

After K-means clustering, SIFTED generated all possible feature combinations by selecting one feature from each cluster. Each feature combination was evaluated by projecting the instances into a temporary 2D space using Principal Component Analysis (PCA), generating a set of 2D coordinates. SIFTED then used the resulting 2D coordinates to train a Random Forest model for each algorithm and compute the out-of-bag (OOB) error. The feature combination minimizing the average out-of-bag (OOB) error across all models was selected, as it defines a better instance space by selecting the most relevant features for algorithm performance. This selection provides the set of features used as input by PILOT to construct the projection matrix.

\subsubsection{PILOT}
PILOT, short for Projecting Instances with Linearly Observable Trends, projects high-dimensional feature spaces into a two-dimensional (2D) space, aiming to create linear relationships between instance features and algorithm performance for improved visual interpretation and pattern identification. Instead of relying on classic dimensionality reduction techniques, PILOT formulates an optimization problem to transform the instances into a 2D space while ensuring linear trends. The optimization aims to minimize the difference between original and projected feature and performance values, calculated based on a linear approximation in the 2D space. Mathematical details can be found in Smith and Muñoz \cite{smith2023instance}. PILOT uses the selected features to construct the projection matrix, such matrix is a key output of ISA. The PILOT method is controlled by the following parameters:

\begin{itemize}
    \item $\mathbf{N_{try}}$: The number of random restarts, with a default value of $\mathbf{N_{try}}=30$;
    \item $\phi_{\text{num}}$: A binary flag,  $\phi_{\text{num}=}$false, selects the analytical solution for the optimization problem rather than the numerical one.
\end{itemize}

%% file: tex05_Results.tex
\section{Instance Space of CVRP benchmarks}
\label{sec_results}

The methodological decisions detailed in Section \ref{sec:methods} culminate in the formulation of Equation \eqref{eqprojectionmatrix}, which defines both a projection matrix and its corresponding feature vector. Consequently, any CVRP instance, characterized by its feature vector, can be projected into a two-dimensional space using this projection matrix. This section focuses on how formulations like Equation \eqref{eqprojectionmatrix} can be employed to offer novel tools for CVRP research. Through this methodology, we demonstrate the potential of the instance space for a deeper comparative analysis than superficial comparisons based solely on basic aggregate average performance.

\begin{figure}[h]
\centerline{\includegraphics[width=0.38\textwidth]{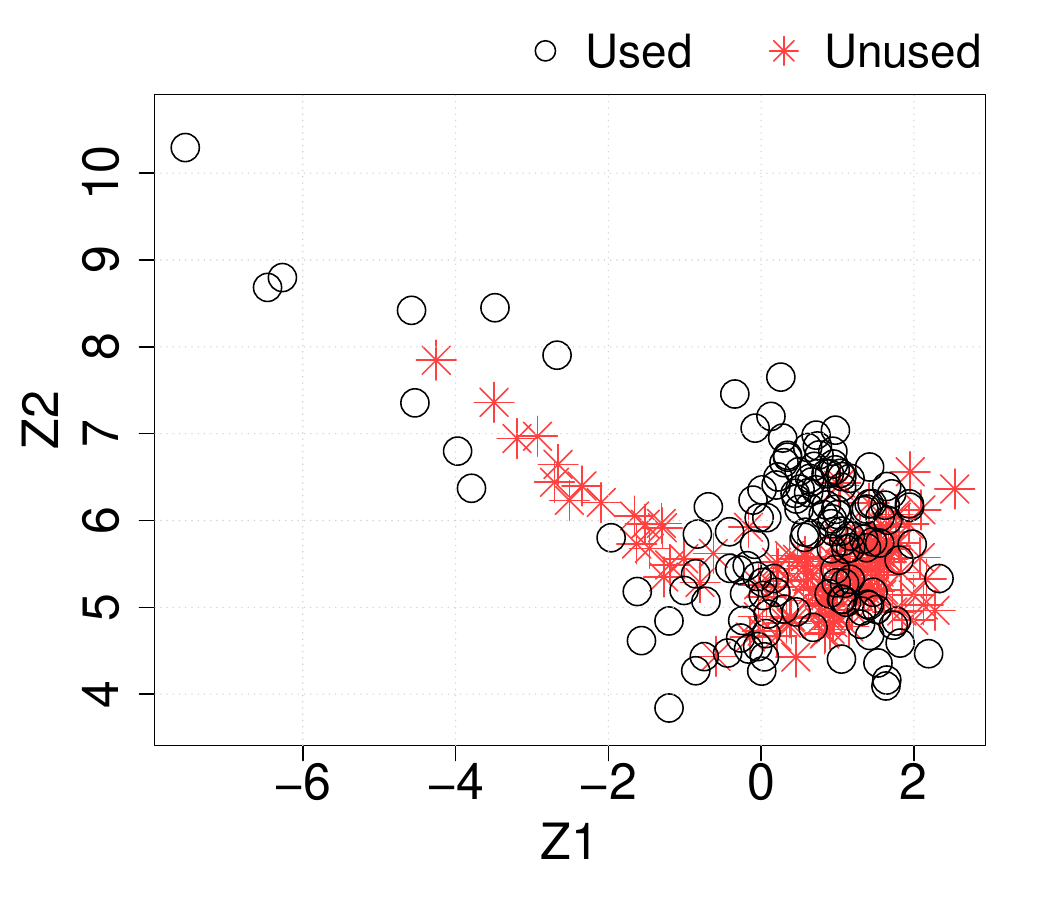}}
\caption{ISA projection (Z1 vs. Z2) of CVRPLib instances. Black circles used in DIMACS and red stars unused.}
\label{fig:cvrplib_projection_all}
\end{figure}

Figure~\ref{fig:cvrplib_projection_all} presents the projected instance space of CVRP instances available in CVRPLib, where black circles denote instances used in the first phase of the competition and red stars represent unused instances. The figure presents a predominantly grouping of instances, with some diverging in an almost linear fashion. From an ISA perspective, this distribution suggests a predominance of specific feature value combinations, indicating that the CVRP literature has not explored a sufficiently diverse range of instances with different characteristics. Furthermore, the same figure allows us to conclude that the organizers selected a good subset of instances, which could be refined by adding some instances from under-represented areas – i.e., red areas without black circles –, and excluding some instances from overpopulated regions – i.e., areas with a high concentration of black circles on the top right.

\noindent
\begin{minipage}{\linewidth}
\vspace{-0.1\baselineskip} % Ajusta o espaço vertical

\begin{equation}
    \label{eqprojectionmatrix}
    \renewcommand\arraystretch{0.4} % Reduz o espaçamento vertical da matriz
    \begin{aligned}
    Z = &
        \begin{split}
            & \displaystyle
            \begin{bmatrix}
               \scriptstyle
                    -0.93 & -0.34 \\
                    -0.29 & ~0.73 \\
                    -0.67 & ~0.62 \\
                    -1.23 & ~0.89 \\
                    -1.07 & ~0.19 \\
                    -0.91 & ~0.80 \\
                    -0.45 & ~0.35 \\
                    -0.58 & ~0.50 \\
                    -0.43 & ~0.96 \\
                    -0.52 & ~1.12 \\
                    -0.65 & ~0.48 \\
                    -0.48 & ~0.86 \\
                    -0.57 & ~0.86 \\
                    ~0.82 & ~0.61 \\
                    ~0.42 & ~1.12 \\
                    ~0.52 & ~0.85 \\
                    -0.48 & ~0.32 \\
                    ~0.56 & ~0.73 \\
                    ~0.61 & ~0.93 \\
                    ~0.11 & ~0.74 \\
                    ~0.51 & ~0.66 \\
                    -0.19 & ~0.36 \\
                    -0.42 & ~1.63
            \end{bmatrix}^{T}
            & 
            \begin{bmatrix}
               \scriptstyle
                \text{NN3~(sd)} \\
                \text{ND8~(var)} \\
                \text{P5~(mean)} \\
                \text{NN3~(skew)} \\
                \text{P6~(sd)} \\
                \text{P4~(mean)} \\
                \text{P11~(skew)} \\
                \text{ND2} \\
                \text{NN2~(max)} \\
                \text{NN2~(skew)} \\
                \text{VRP4} \\
                \text{P10~(mean)} \\
                \text{MST3~(median)} \\
                \text{ND5~(mean)} \\
                \text{P7~(var)} \\
                \text{P2~(mean)} \\
                \text{P1~(mean)} \\
                \text{P3~(mean)} \\
                \text{G2} \\
                \text{P6~(skew)} \\
                \text{P9~(mean)} \\
                \text{P5~(skew)} \\
                \text{MST2~(mean)}
              \end{bmatrix}
          \end{split}
    \end{aligned}
\end{equation}

% --- Início da "Legenda" ---
\begin{center}
    \parbox{0.9\textwidth}{\small \textbf{Eq. \eqref{eqprojectionmatrix}:}  The 2D projection matrix Z derived from the PILOT stage, along with the 23 input features selected by SIFTED.}
\end{center}
% --- Fim da "Legenda" ---

\vspace{-0.1\baselineskip} % Ajuste o espaço vertical após a equação
\end{minipage}

\begin{figure*}[t] % 'figure*' para ocupar a largura total
    \centering
    \begin{tabular}{ccc}
        \subfloat[Set E \cite{christofides1969algorithm}]{\includegraphics[width=0.25\textwidth]{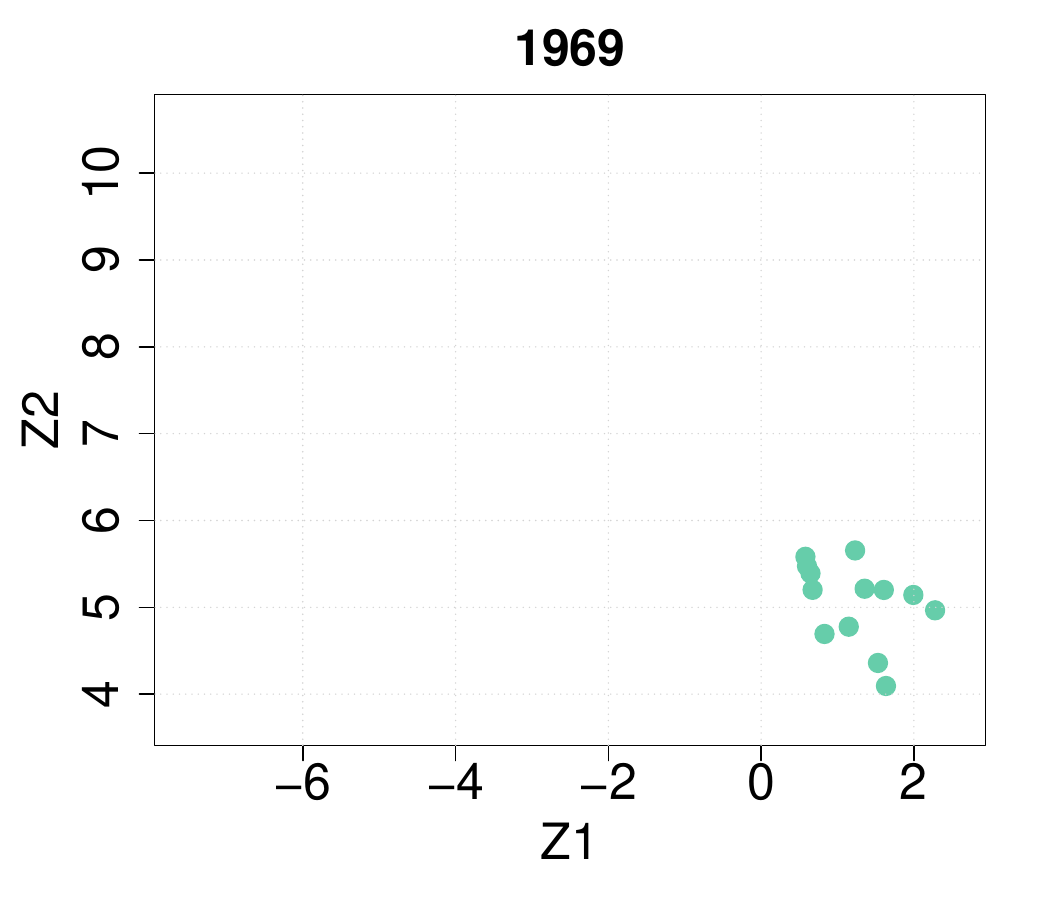}\label{fig:family_1}} &
        \subfloat[Sets M and CMT]{\includegraphics[width=0.25\textwidth]{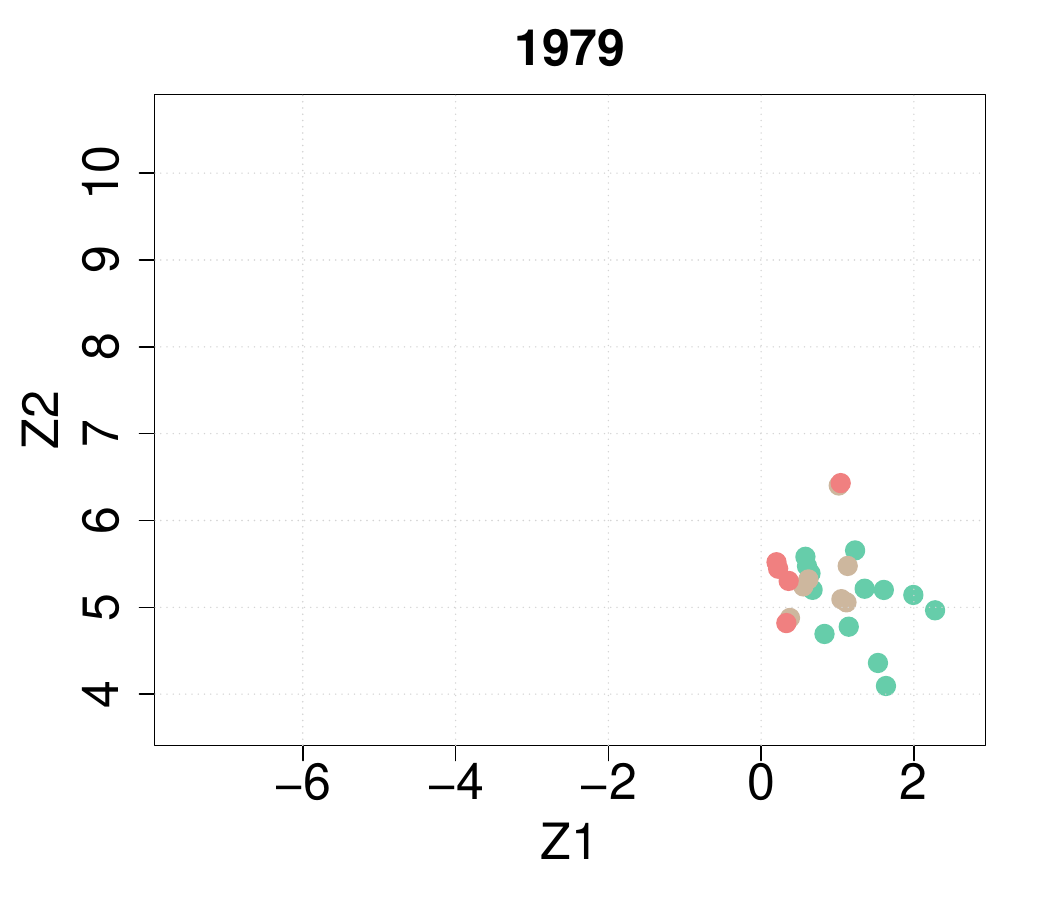}\label{fig:family_2}} &
        \subfloat[Set F]{\includegraphics[width=0.25\textwidth]{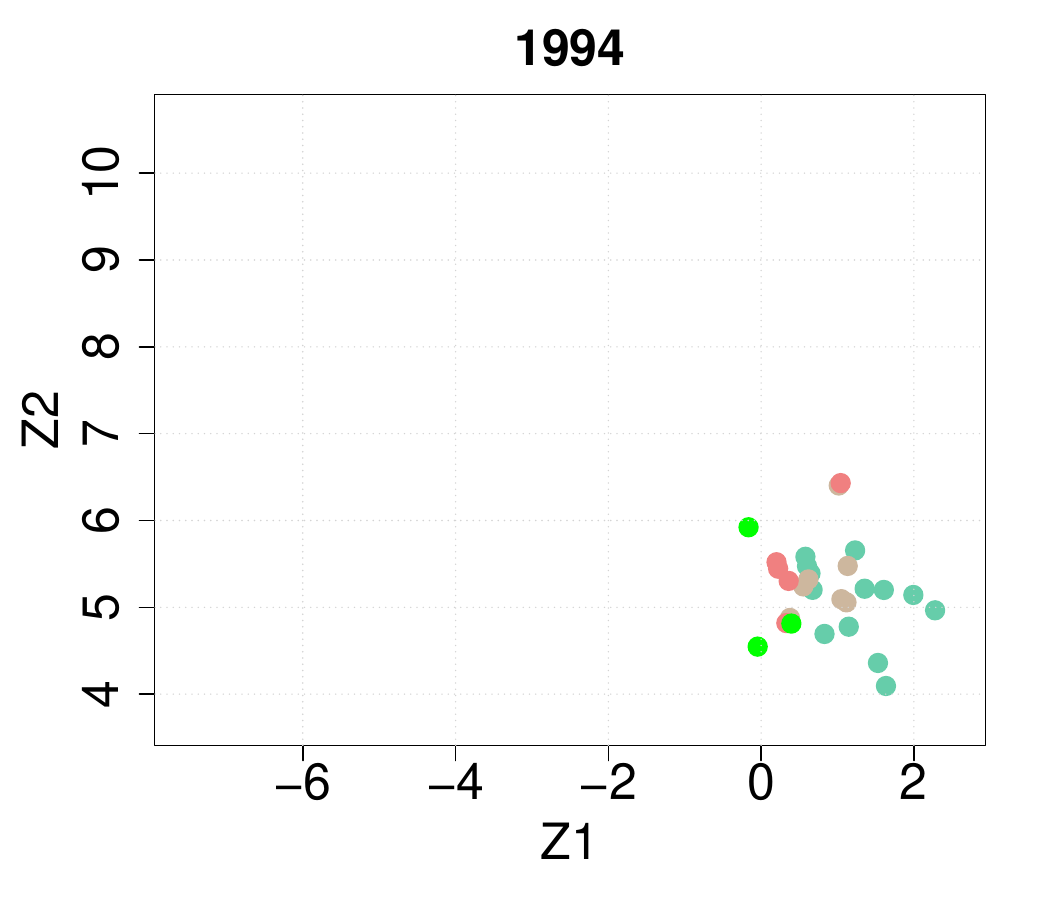}\label{fig:family_3}} \\
        \medskip
        \subfloat[Sets A, B and P]{\includegraphics[width=0.25\textwidth]{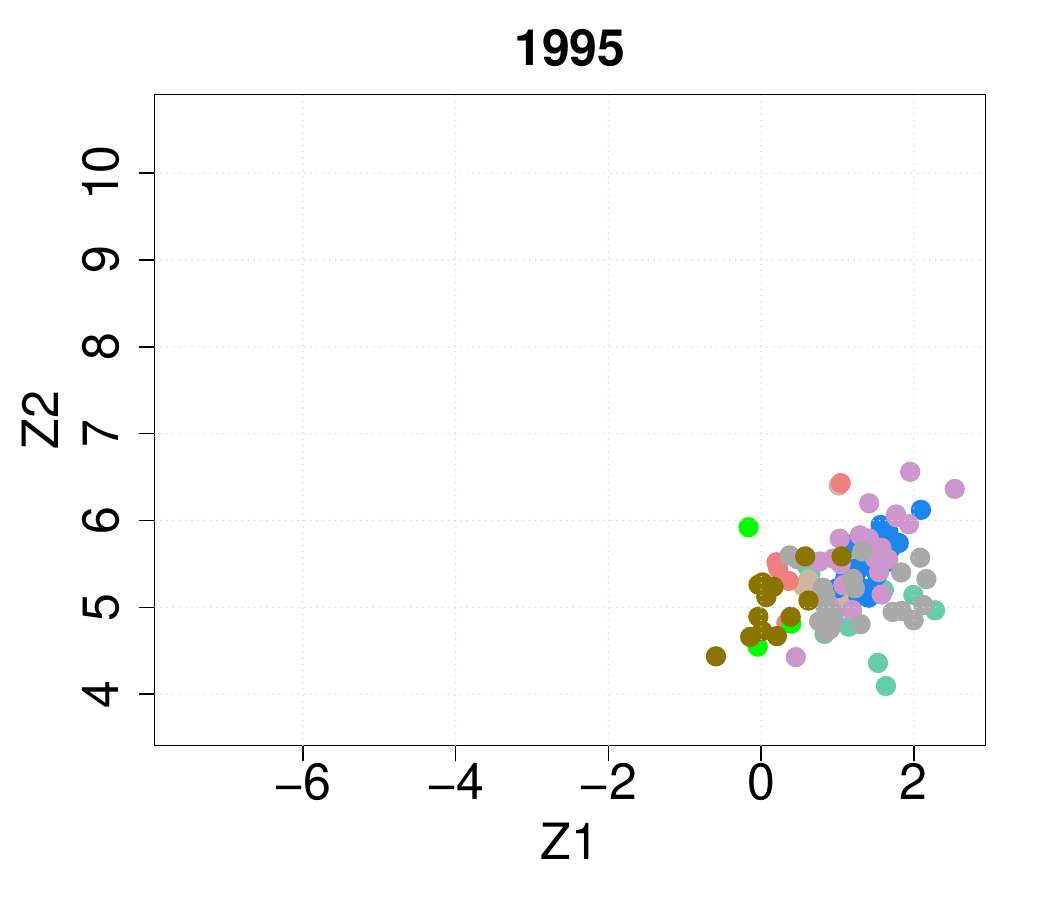}\label{fig:family_4}} &
        \subfloat[Golden instances]{\includegraphics[width=0.25\textwidth]{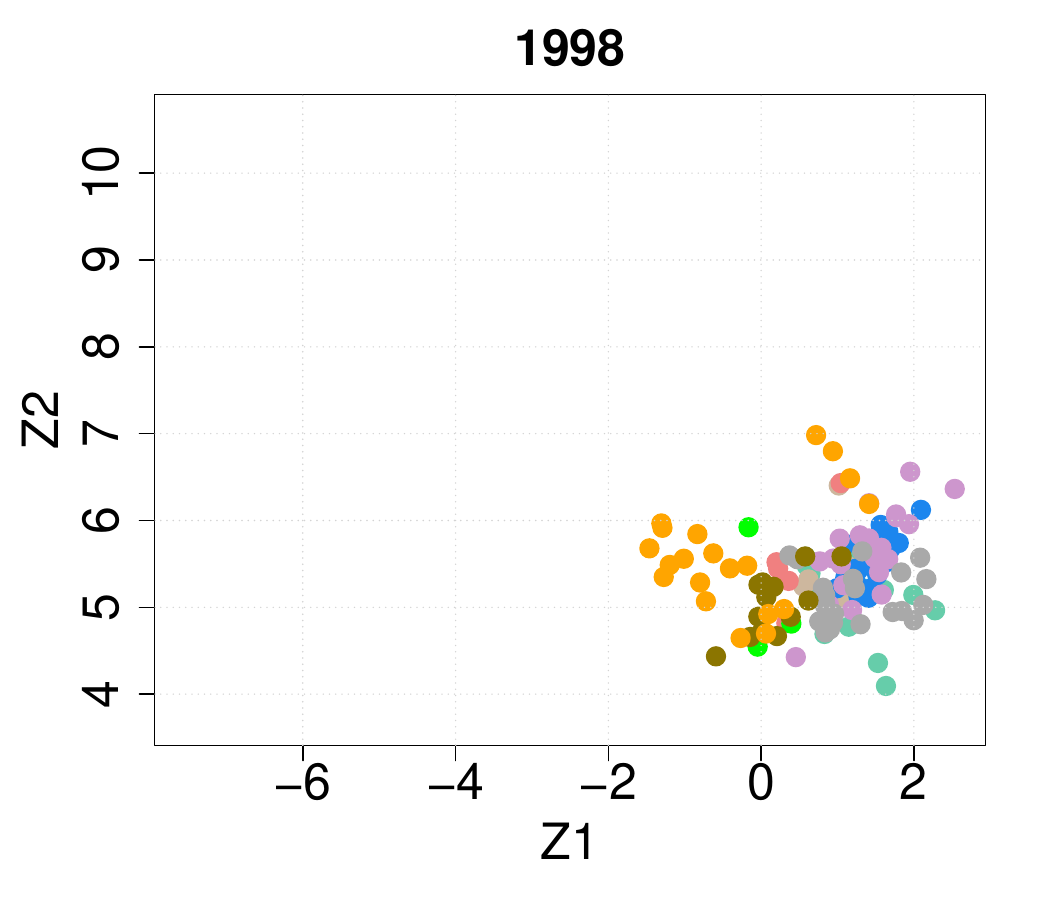}\label{fig:family_5}} &
        \subfloat[Li instances]{\includegraphics[width=0.25\textwidth]{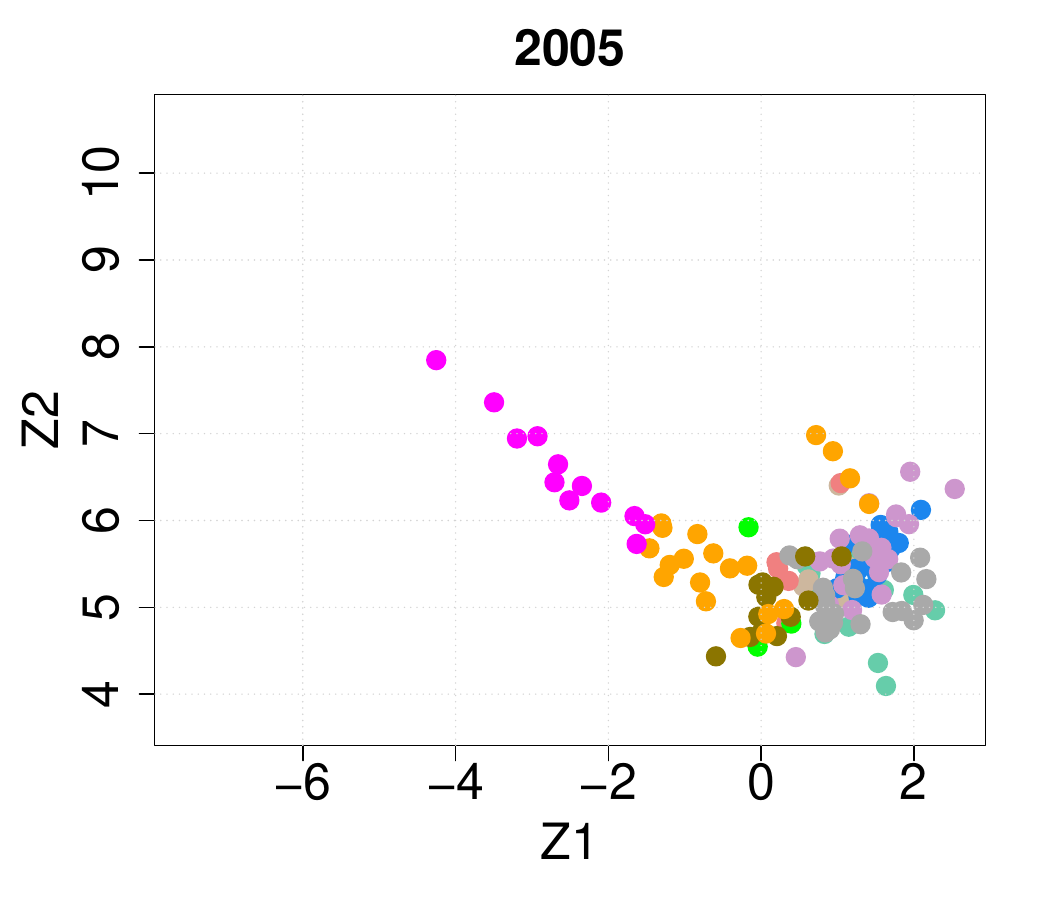}\label{fig:family_6}} \\
        \medskip
        \subfloat[X]{\includegraphics[width=0.25\textwidth]{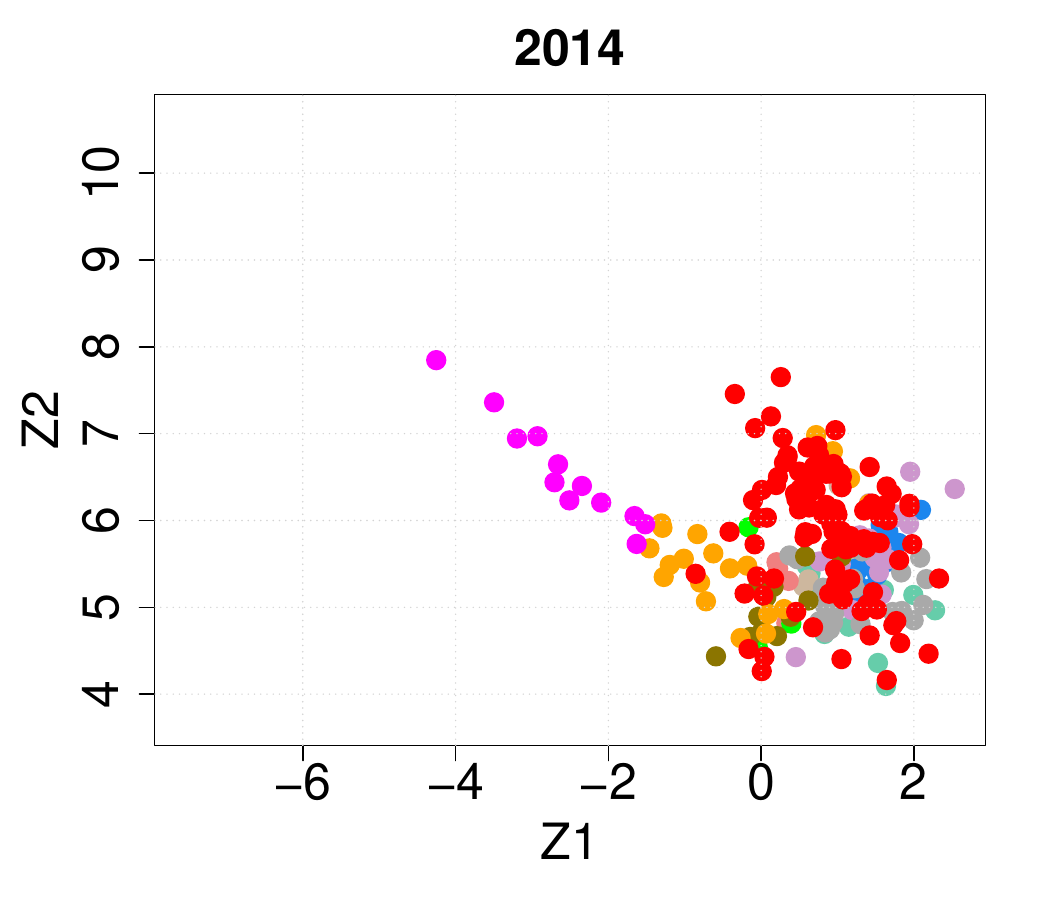}\label{fig:family_7}} &
        \subfloat[AGS]{\includegraphics[width=0.25\textwidth]{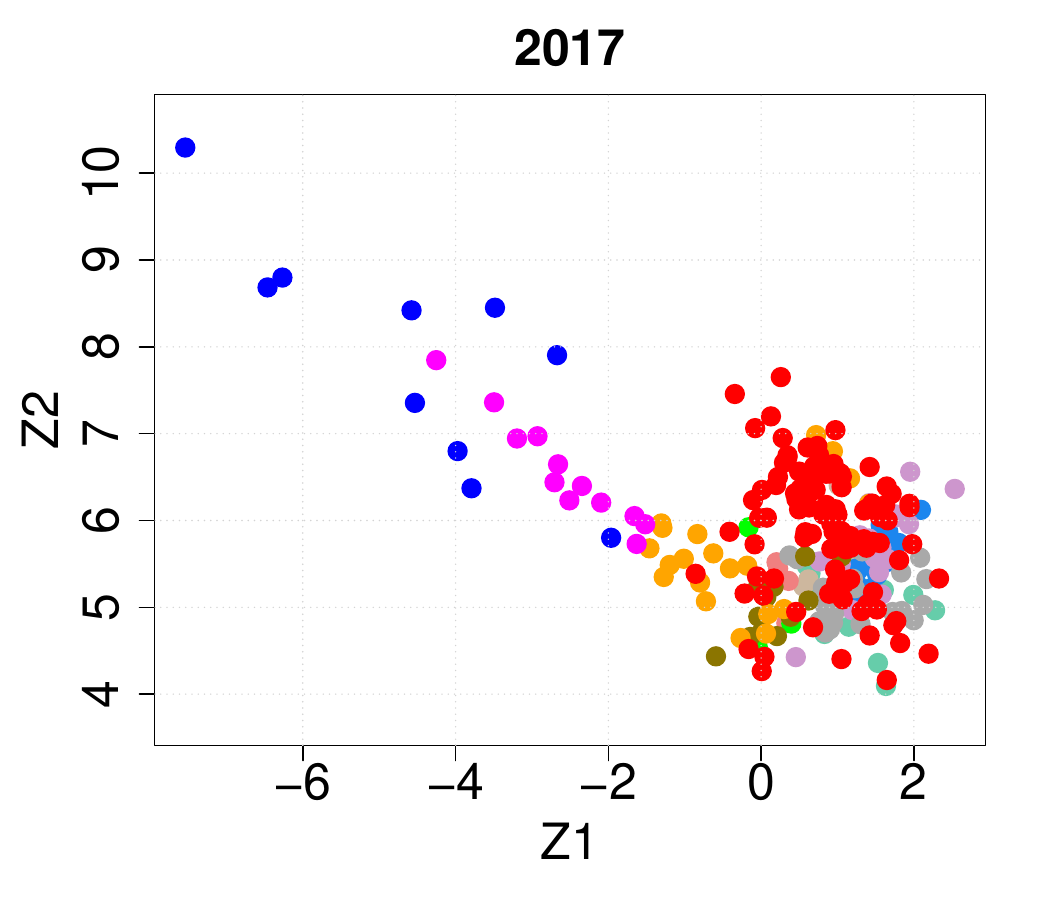}\label{fig:family_8}} &
        \subfloat[DIMACS]{\includegraphics[width=0.25\textwidth]{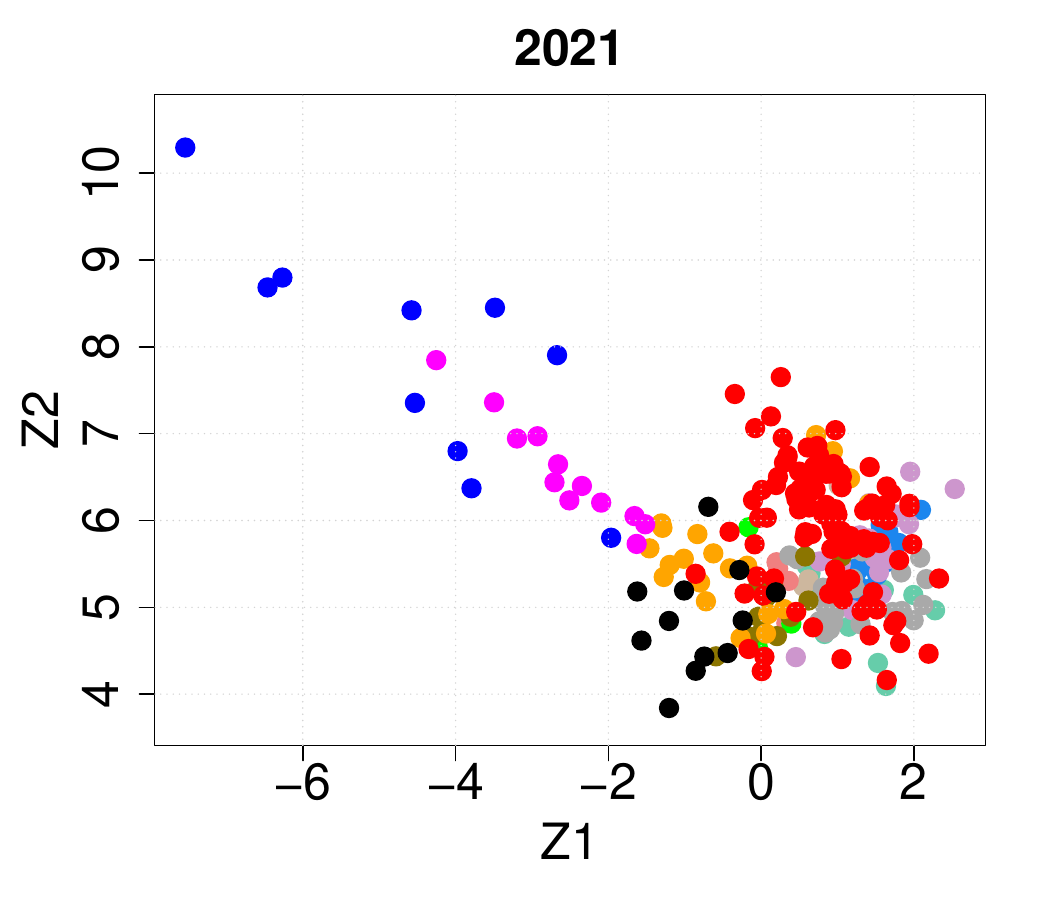}\label{fig:family_9}}
        \cr
        \subfloat[Color Key]{\includegraphics[width=0.25\textwidth]{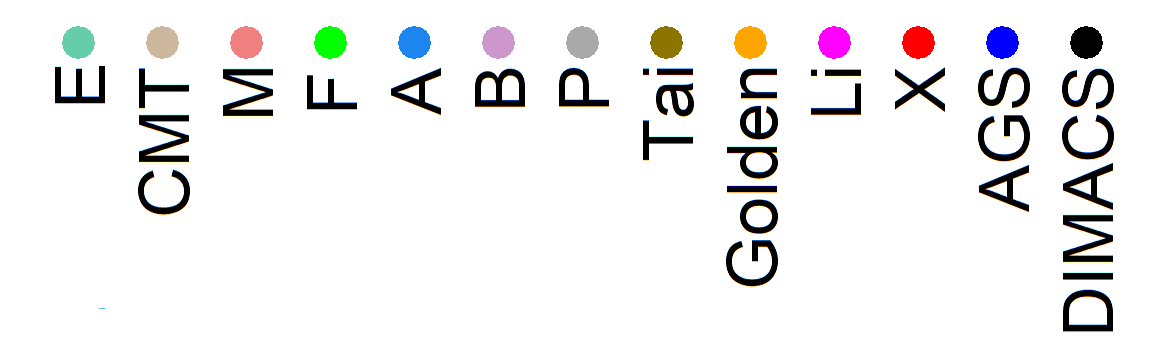}\label{fig:family_LEGEND}}
    \end{tabular}
    \caption{A time-based visualization of the CVRP instance space, with different colors representing different sets of instances proposed over the years, showing how the space was populated over time.}
    \label{fig:cvrpOverYears}
\end{figure*}

It is widely recognized within the CVRP literature that benchmarks have often been created and shared by the works of different research groups. Equation \eqref{eqprojectionmatrix} allows us to observe and analyze the evolution of CVRP instances over the years. By comparing Figure \ref{fig:family_1} to Figure \ref{fig:family_9}, we can observe how the instance space has been populated over time by different instance sets, where each instance set is represented by a different color. Figure~\ref{fig:family_1} shows the earliest set of CVRP instances proposed by Christofides and Eilon \cite{christofides1969algorithm} in 1969, known as Set E. Ten years later, Christofides, Mingozzi, and Toth \cite{christofides1979vehicle} introduced Sets M and CMT; Figure~\ref{fig:family_2} reveals that these new sets present a very similar feature distribution as Set E. Fifteen years later, Fisher \cite{fisher1994optimal} proposed three new instances, which we refer to as Set F, and they can be seen in Figure \ref{fig:family_3}. One year later, Augerat et al \cite{augerat1995computational} introduced Sets A, B, and P as it is possible to see in Figure \ref{fig:family_4} which made the bottom right of the instance space more densely populated.

\begin{figure*}[h!]
    \centering
    \begin{tabular}{ccc}
        \subfloat[Winner]{\includegraphics[width=0.25\textwidth]{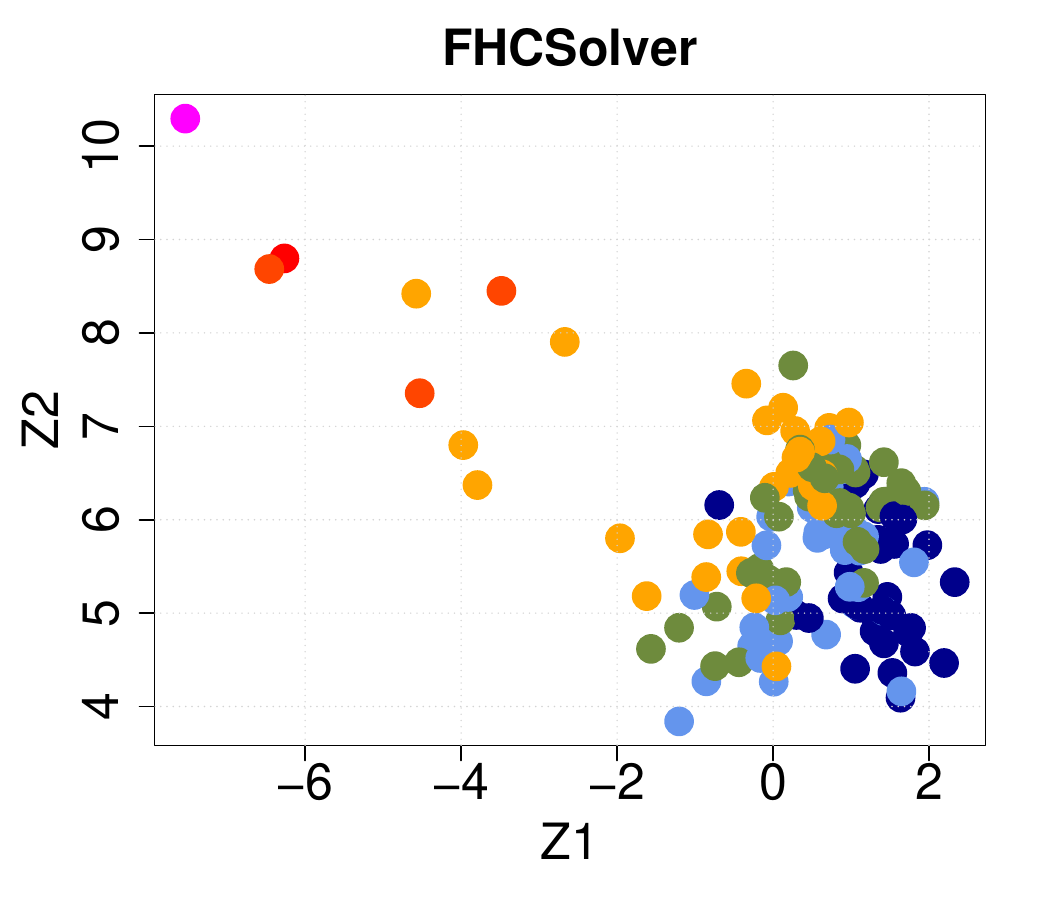}\label{fig:FHCSolver}} 
        &
        \subfloat[Fourth place]{\includegraphics[width=0.25\textwidth]{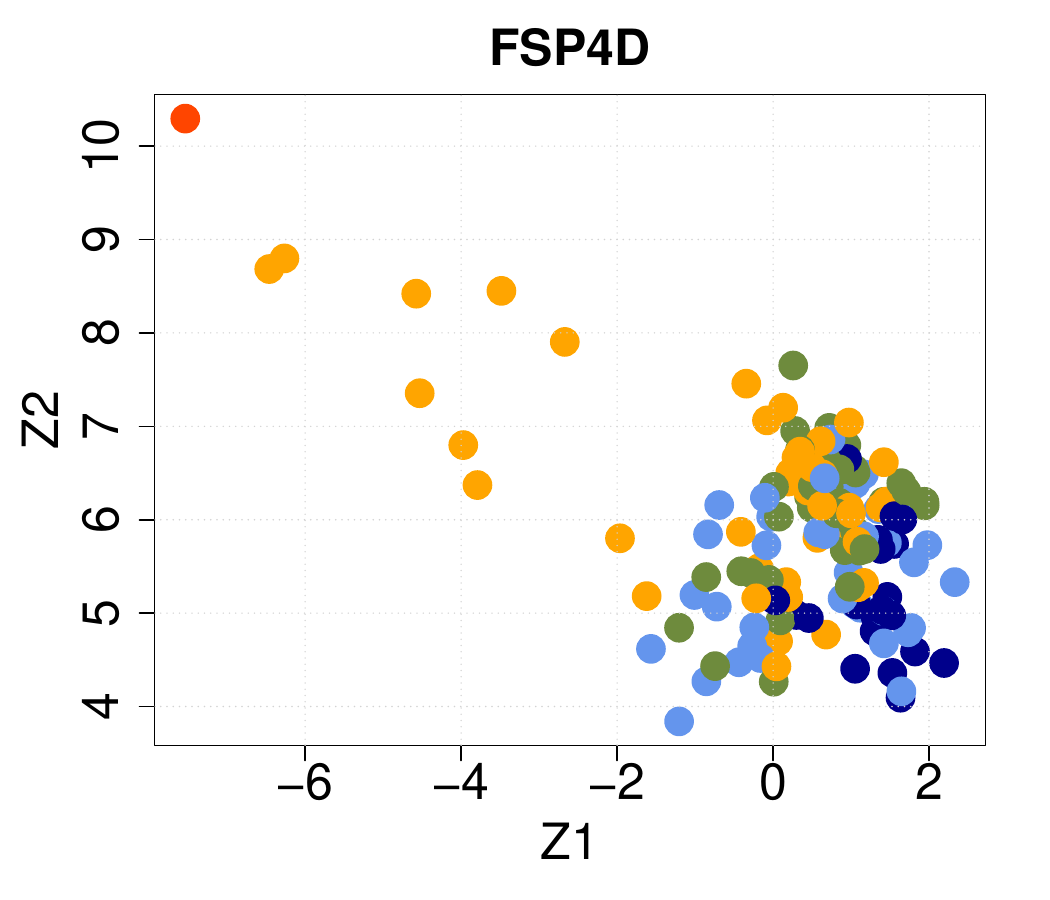}\label{fig:FSP4D}}
        &
        \subfloat[Eighth place]{\includegraphics[width=0.25\textwidth]{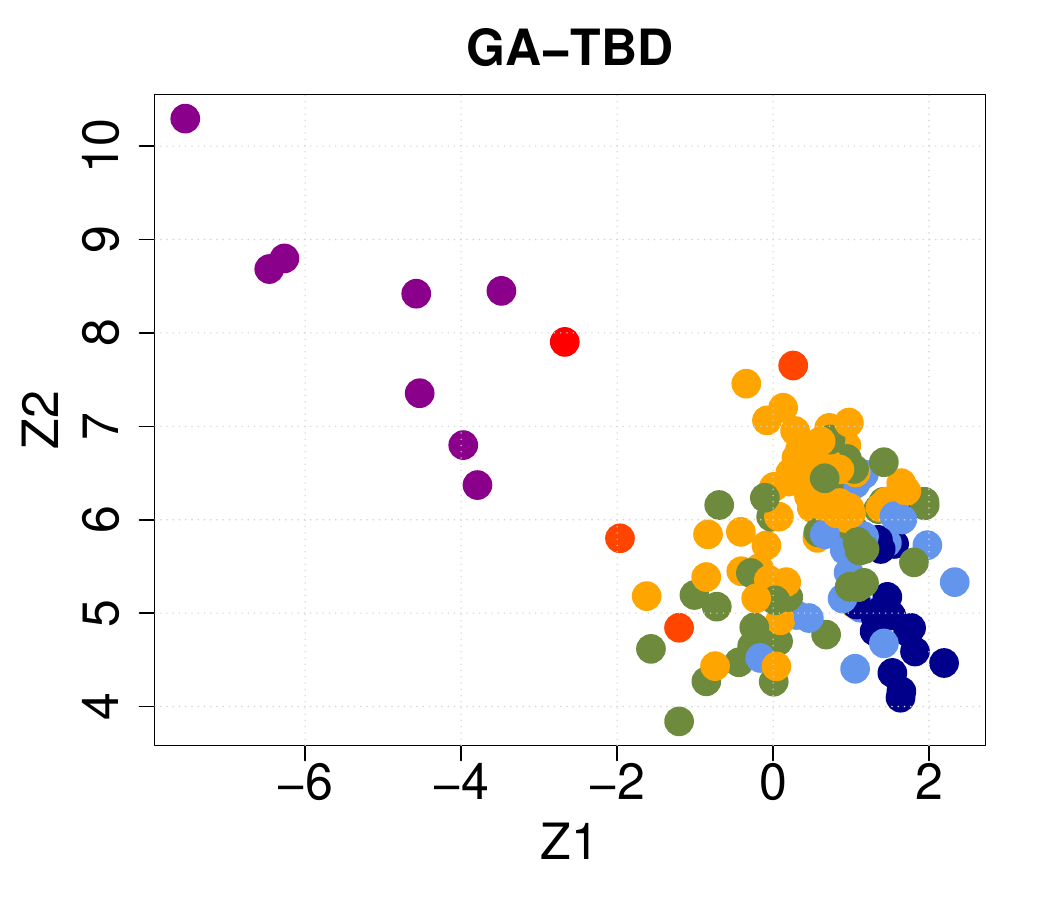}\label{fig:GATBD}} 
    \end{tabular}
    \caption{Performance of MHs FHCSolver, FSP4D, and GATBD from left to right.}
    \label{fig:top8metaheuristic}
\end{figure*}

\begin{figure}[h]
\centerline{\includegraphics[width=0.3\textwidth]{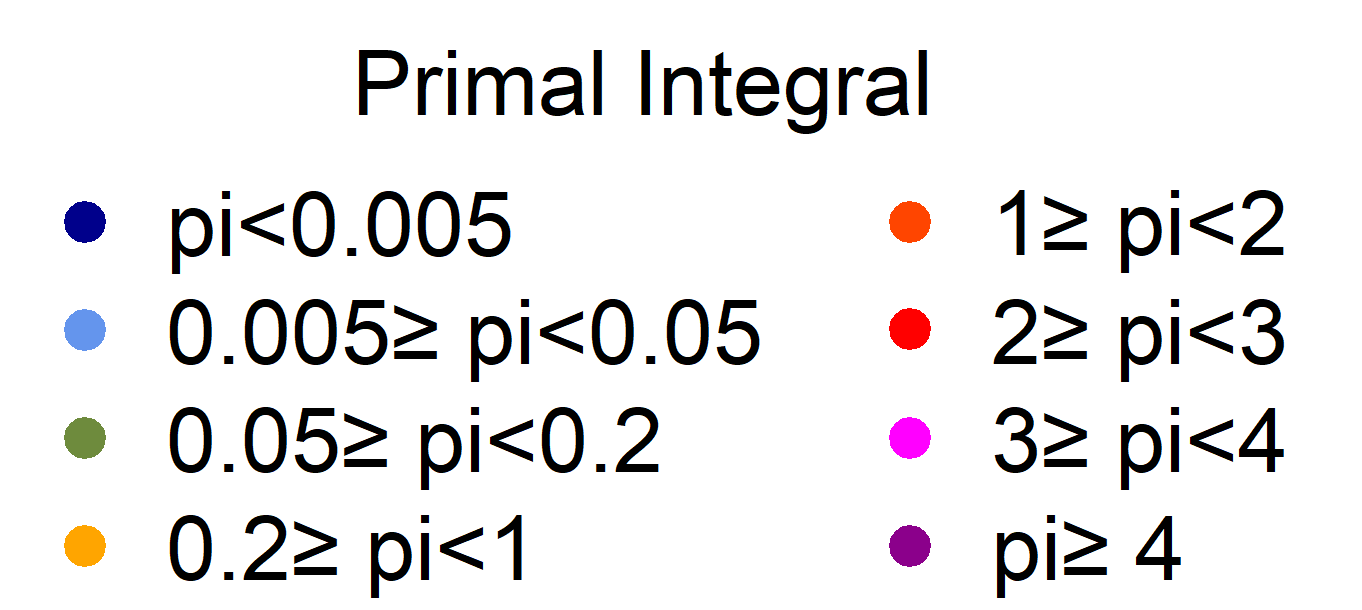}}
\caption{The color scale used to interpret MHs performance in the instance space visualizations. Different colors indicate different performance levels, measured by the Primal Integral.}
\label{fig:legAlgoPer}
\end{figure}

To this point, most CVRP instances, with at most two hundred customers, were designed primarily to showcase the power of new solution algorithms. However, they were no longer useful benchmarks after Fukasawa et al. \cite{fukasawa2006robust} developed an algorithm able to solve them in 2006, except just three. Subsequently, the scientific community was able to address those remaining issues with the work of Ropke \cite{ropke2012branching} (2012), Contardo and Martinelli \cite{contardo2014new} (2014), and Pecin et al. \cite{pecin2017improved} (2017), who proposed algorithms that solved the last unsolved benchmarks (M-n151-k12, M-n200-k16 and M-n200-k17). As a consequence, the literature reveals that researchers started working to propose instances that were more difficult to solve. Figures \ref{fig:family_5} to \ref{fig:family_9} illustrate this shift in research focus by showing new instances occupying previously unexplored areas.

Figure \ref{fig:family_5} shows the Golden set proposed in 1998 by Golden et al. \cite{golden1998impact}, who compiled instances from various sources, including scenarios with 240 to 483 customers and emphasized the need for diverse datasets in comparative studies. Note that the Golden set is concentrated in two regions, mostly populating previously underrepresented areas. Seven years later, Li \cite{li2005very} proposed new instances, advocating new challenges in VRP research, and those new instances are displayed as pink dots in Figure \ref{fig:family_6}. Subsequently, in 2017, Uchoa et al. \cite{uchoa2017new} proposed a new class of instances, denoted by X, based on the observation that the Golden instances represent artificial formulations when contrasted with real-world applications. Figure \ref{fig:family_7} reveals that the set X fills the previously observed white space between Golden instances – that is, explores combinations of instance characteristics not previously represented – extends the instance space to the right, and also creates more densely populated regions. Building on this line of reasoning, in 2019, Arnold et al. \cite{arnold2019efficiently}, following Li's argument, highlighted the limitations of the most popular benchmarks (Golden et al., 1998; Uchoa et al., 2017) for having a limited number of customers compared to real-world problems. Figure \ref{fig:family_8} shows how Arnold et al. proposed a new set of instances that expands the instance space to the top-left. Finally, Figure \ref{fig:family_9} shows the real-world instances, represented by black dots, offered in the DIMACS Challenge. This set, referred to as the DIMACS set, broadens the instance space to the bottom-right.

Given the history of how instances were proposed, state-of-the-art algorithms are expected to perform better in the predominantly dense region ($Z_1>-2$ and $Z_2$<6) than in other areas, since the initial benchmarks were concentrated there. This dense region, primarily composed of instances from the earlier sets ABEFMP, represents the area of the instance space where most MHs have been trained over the years. As a result, the MHs might be overly tuned to the specific characteristics of these ABEFMP instances. Figures \ref{fig:FHCSolver}, \ref{fig:FSP4D}, and \ref{fig:GATBD} show the performance of the first, fourth, and eighth place finishers, respectively. Note that these figures highlight a concentration of blue data points in the dense region, indicating good performance of the MHs (see the color scale in Figure \ref{fig:legAlgoPer}, where blue shades represent better performance). However, a closer examination of the less dense regions of the instance space reveals a more nuanced picture of algorithm performance. When considering the less dense regions of the instance space, importantly, the instance projections provide a visual tool that can generate insights that go beyond those provided by a simple median performance analysis. Observe that in a hypothetical scenario with instances that follow a nearly linear distribution of key characteristics, the ranking of the competitors would change. When comparing the results from FHCSolver (Figure \ref{fig:FHCSolver}) and FSP4D (Figure \ref{fig:FSP4D}), it is clear that FSP4D would achieve a higher ranking if only these regions were considered.

Although projection is a useful tool to analyze instances and to interpret algorithm performance, it is not straightforward to understand the reasons behind why each instance is located where they are. The multi-causality and interdependence of the method are the main drivers for such a challenge, in which the position of each data point is typically influenced by multiple causes that present interrelations. We put some effort into an attempt to shed light on which characteristics make the instance be positioned in a given area. To do so, we analyzed the features (the set F) through clustering techniques (K-means and DBSCAN) and decision trees. However, the results were not remarkable. Nonetheless, it is relevant that we found a moderate correlation, measured by Pearson's correlation coefficient, between the number of customers and axis $Z_1$ (r=-0.6) and $Z_2$ (r=0.51).

%% file: tex06_conclusion.tex
\section{Conclusion and Future Works}
\label{sec:conclusion}

This paper focused on the challenge of understanding the nuanced relationships between instance characteristics and metaheuristic (MH) performance, a key issue for advancing the state-of-the-art in CVRP research. We successfully demonstrate the use of Instance Space Analysis (ISA) as a novel perspective on this key issue.  In our analyses we used the data provided by the DIMACS 12th Implementation Challenge on Vehicle Routing provided. Through the PRELIM, SIFTED, and PILOT stages, we were able to identify twenty three  relevant instance features and propose a projection matrix. Such matrix is the key contribution of our work, since it  enables the straightforward incorporation of new instances providing a new method for instance analysis in the CVRP field.

Our work presents a visual approach to analyze the relationship between CVRP instances and MH performance, revealing that the performance of state-of-the-art algorithms reflects the historical order in which instances were proposed, with newer instances posing a persistent challenge. While we also addressed the challenge of understanding which features determine an instance's position in the projected space (i.e., what explains the proximity or distance between data points), this proved to be a complex task that warrants further investigation. Understanding the interrelations of the identified features is essential to guide the development of more efficient MHs and to generate challenging instances for different regions of the instance space.